\documentclass[letterpaper, 10 pt, conference]{ieeeconf}  

\IEEEoverridecommandlockouts                              

\usepackage{cite}
\usepackage{hyperref}
\usepackage{amsmath,amssymb,amsfonts}
\usepackage{algorithmic}
\usepackage{graphicx}
\usepackage{booktabs} 
\usepackage{textcomp}
\usepackage{xcolor}
\usepackage{verbatim} 
\usepackage[linesnumbered,lined,commentsnumbered,ruled]{algorithm2e}
\usepackage{adjustbox}
\usepackage{tikz}

\usepackage{optidef} 

\newcommand{\todo}[1]{{\color{red!60!black} TODO:~#1}}%
\newcommand{\ub}[1]{\overline{ #1 }}
\newcommand{\lb}[1]{\underline{ #1 }}

\newcommand{\R}{\mathbb{R}}
\newcommand{\Id}{\mathbb{I}}

\newcommand{\norm}[1]{\left\lVert #1 \right\rVert}

\newcommand{\mpcdef}{{Parameterized Model Predictive Planner}}
\newcommand{\mpc}{{MPP}}
\newcommand{\mpcfull}{{Parameterized Model Predictive Planner}}
\newcommand{\rl}{{high-level RL policy}}
\newcommand{\mpcrldef}{{\emph{Hierarchical Learning-based Predictive Planner}}}
\newcommand{\mpcrl}{{HILEPP}}

\newcommand{\state}{x}
\newcommand{\lon}{\zeta}
\newcommand{\controls}{u}

\DeclareMathSymbol{\shortminus}{\mathbin}{AMSa}{"39}

\newcommand{\ra}[1]{\renewcommand{\arraystretch}{#1}}

\newcommand\copyrighttext{%
	\footnotesize \textcopyright 2022 IEEE. Personal use of this material is permitted. Permission from IEEE must be obtained for all other uses, in any current or future media, including reprinting/republishing this material for advertising or promotional purposes, creating new collective works, for resale or redistribution to servers or lists, or reuse of any copyrighted component of this work in other works.}
\newcommand\copyrightnotice{%
	\begin{tikzpicture}[remember picture,overlay]
		\node[anchor=south,yshift=5pt] at (current page.south) {\fbox{\parbox{\dimexpr\textwidth-\fboxsep-\fboxrule\relax}{\copyrighttext}}};
	\end{tikzpicture}%
}

\title{\LARGE \bf
	A Hierarchical Approach for Strategic Motion Planning in Autonomous Racing
}

\author{Rudolf Reiter$^{1}$, Jasper Hoffmann$^{2}$, Joschka Boedecker$^{2}$ and Moritz Diehl$^{1,3}$
	\thanks{$^{1}$Department of Microsystems Engineering, University Freiburg, 79110 Freiburg, Germany
		{\tt\small \{rudolf.reiter, moritz.diehl\}@imtek.uni-freiburg.de}}%
	\thanks{$^{2}$Neurorobotics Lab, University Freiburg, 79110 Freiburg, Germany 
		{\tt\small \{hoffmaja, jboedeck\}@informatik.uni-freiburg.de}}%
	\thanks{$^{3}$Department of Mathematics, University Freiburg, 79110 Freiburg, Germany
	}%
}

\begin{document}
	
\maketitle
\copyrightnotice
\thispagestyle{empty}
\pagestyle{empty}

\begin{abstract}
	We present an approach for safe trajectory planning, where a strategic task related to autonomous racing is learned sample-efficient within a simulation environment.	 
	A high-level policy, represented as a neural network, outputs a reward specification that is used within the cost function of a parametric nonlinear model predictive controller (NMPC).
	By including constraints and vehicle kinematics in the NLP, we are able to guarantee safe and feasible trajectories related to the used model.
	Compared to classical reinforcement learning (RL), our approach restricts the exploration to safe trajectories, starts with a good prior performance and yields full trajectories that can be passed to a tracking lowest-level controller.
	We do not address the lowest-level controller in this work and assume perfect tracking of feasible trajectories.
	We show the superior performance of our algorithm on simulated racing tasks that include high-level decision making. 
	The vehicle learns to efficiently overtake slower vehicles and to avoid getting overtaken by blocking faster vehicles.
\end{abstract}
\section{Introduction}
	Motion planning for autonomous racing is challenging due to the fact that vehicles operate at the performance limits.
	Furthermore, planning requires interactive and strategic, yet safe behavior.
    In this work we focus on strategic planning for fixed opponent policies with safety guarantees.
	Current research usually is based on either graph-based, sampling-based, learning-based or optimization-based planners \cite{betz_2022_1, Paden2016}.
	We propose a combination of model-predictive control (MPC) and a neural network (NN), trained by a reinforcement learning (RL) algorithm within simulation.
	MPC is a powerful  optimization-based technique that is used commonly for solving trajectory planning and control problems.
	The use of efficient numerical solvers and the possibility to incorporate constraints directly \cite{Rawlings2017} makes MPC attractive in terms of safety, explainability and performance.
	Nevertheless, in problems like interactive driving, it is difficult to model the behavior of other vehicles.
	In contrast to MPC, RL is an exploration-driven approach for solving optimal control problems. Instead of an optimization friendly model, RL only requires samples of the dynamics and can in theory optimize over arbitrary cost functions.
	The flexibility of RL comes at the cost of a high sample inefficiency that is often unfavorable for real-world applications, where data is expensive and rare.
	Furthermore, RL in the general setting lacks safety-guarantees.
    However, once the amount and quality of data is sufficient, the learned policies can show impressive results \cite{granTurismo2022}.
	In this paper, we combine MPC and RL by using an MPC-inspired low-level trajectory planner to yield kinematic feasible and safe trajectories and use the {\rl} for  strategic decision-making.
	We use the expression {\mpc} ({\mpcdef}) to refer to a MPC-based planner, which outputs feasible reference trajectories that we assume to be tracked by a \emph{lowest}-level control systems. 
	This hierarchical approach is common in automotive software stacks \cite{Vazquez2020, Paden2016}.
	We use the {\mpc} to formulate safety-critical constraints and basic time-optimal behavior but let the cost function be subject to changes by the {\rl}. 
	Particularly, we propose an interface where the {\rl} outputs a reference in the Frenet coordinate frame.
	With this approach, we start with a good prior strategy for known model parts and can guarantee safe behavior with respect to the chosen vehicle model and the prediction of the opponents.
	The structure of this paper is as follows.
	In Sec.~\ref{sec:general method}, we motivate our approach by a similar formulation named \emph{safety filter} \cite{wabersich2021predictive}, in Sec.~\ref{sec:specific_method} we describe the MPC-based planner and in Sec.~\ref{sec:rl_policy} we explain the implementation of the {\rl} and how we train it.
	Finally, in Sec.~\ref{sec:results} we evaluate the algorithm, which we refer to as {\mpcrl} ({\mpcrldef}), in a multi-agent simulation which involves strategic decision making.\\
	\textit{Contribution}: We contribute by a derivation and evaluation of a sample-efficient and safe motion planning algorithm for autonomous race-cars. 
	It includes a novel cost function formulation for the interaction of MPC and RL with a strong prior performance, real-time applicability and high interpretability.\\
	\textit{Related work}:
	Several works consider RL as a set-point generator for MPC for autonomous agents \cite{greatwood_reinforcement_2019, britoWhereGoNext}.
	As opposed to our approach, they focus on final target-points. 
	Another research branch focuses on safety verification with a so-called "safety-filter" \cite{brunke_safe_2021}. 
	For instance, in \cite{wabersich2021predictive}, a rudimentary MPC variant is proposed that considers constraints by using MPC as a verification module. 
	Similarly, the authors of \cite{lubars2021combining} use MPC as a correction of an RL policy, if a collision check fails.
	RL is also used with the intention of MPC weight tuning, such as in \cite{scaramuzza2020} for UAVs and in \cite{ZarroukiWeights2020} for adaptive control in autonomous driving.
	Related research for motion planning of autonomous racing was recently surveyed in \cite{betz_2022_1}.
	Several works focus on local planning without strategic considerations \cite{Vazquez2020,granTurismo2022}, thus can not directly be used in multi-agent settings. Other works use a game-theoretic framework \cite{liniger_2017} which, however, often limits the applicability due to its complexity.
	An algorithm for obtaining Nash equilibria is iterated best response (IBR), as shown for drone racing in \cite{spica_2018} or for vehicle racing in \cite{wang_2021}. However, IBR has high computation times. An algorithm aiming at the necessary KKT conditions of the generalized Nash equilibrium problem is presented in \cite{cleac_2021}.
	However, the resulting optimization problem is hard to solve. 
	In \cite{schwartnig_2021}, long term strategic behavior is learned through simulation without safety considerations.
\section{Background and Motivation}
\label{sec:general method}
	A trained neural network (NN) as function approximation for the policy~$\pi^\theta(s)$, where~$\theta\in \R^{n_\theta}$ is the learned parameter vector and~$s\in \R^{n_s}$ is the RL environment state, can generally not guarantee safety.
	Safety is related to constraints for states and controls that have to be satisfied at all times.
	Therefore, the authors in \cite{wabersich2021predictive} propose an MPC-based policy~${\pi^\mathrm{S}:\R^{n_a}\rightarrow \R^{n_a}}$ that projects the NN output~$a\in \R^{n_a}$ to a safe control~$u^\mathrm{S}=\pi^\mathrm{S}(x,a)$, where it is guaranteed that~$u^\mathrm{S}\in\mathcal{U}^\mathrm{S}\subseteq\R^{n_a}$.
	The safe set~$\mathcal{U}^\mathrm{S}$ is defined for a known (simple) system model~$\dot{x}=f(x,u)$ with states~$x$ and controls~$u$ and corresponding, often tightened, constraints.
	In this formulation, the input~$u$ has the same interpretation as the action~$a$ and the state~$x$ relates to the model inside the filter.
	Constraint satisfaction for states is expressed via the set membership~${x\in\mathcal{X}}$ and for controls via ${u\in\mathcal{U}}$. 
	The system model is usually transformed to discrete time via an integration function $x_{i+1}= F(x_i,u_i)$ with step size~$\Delta t$.
	When using direct multiple shooting \cite{Bock1984} one obtains decision variables for the state $X=[x_0,\ldots,x_N]\in\R^{n_x \times (N+1)}$ and for the controls $u=[u_0,\ldots,u_{N-1}]\in\R^{n_u \times N}$.
	Since the optimization problem can only be formulated for a finite horizon, a control invariant terminal set~$S^t$, needs to be included.
	The safety filter solves the following optimization problem
	\begin{mini}
		{X,U}			
		{\norm{u_0-\bar{a}}_{R}^2 }
		{\label{eq:safety_filter}} 
		{} 
		\addConstraint{x_0}{= \bar{x}_0,\quad x_N \in \mathcal{S}^\mathrm{t}}{}
		\addConstraint{x_{i+1}}{= F(x_i,u_i),}{\quad i=0,\ldots,N-1}
		\addConstraint{x_{i}}{\in \mathcal{X},u_{i}\in \mathcal{U},}{\quad i=0,\ldots,N-1}
	\end{mini}
	and takes the first control~$u_0^*$ of the solution~$(X^*,U^*)$ as output~$u^\mathrm{S}:=u^*_0$.
	The authors in \cite{wabersich2021predictive} use the filter as a post-processing safety adaption, while we propose to use this formulation as a basis for an online filter, even during learning, which makes it applicable for safety relevant environments.
	We do not require the same physical inputs to our filter, rather modifications to a more general optimization problem, similar to \cite{zanon2020}.
	We propose a more general interface between the {\rl} and MPC, namely a more general cost function $L(X,U,a)$, modified by action $a$.
	Our general version of the {\mpc} as fundamental part of the algorithm solves the optimization problem
	\begin{mini}
		{X,U}
		{L(X,U,a)}			
		{\label{eq:leremo_filter}} 
		{} 
		\addConstraint{x_0}{= \hat{x}_0,\quad x_N \in \mathcal{S}^\mathrm{t}}{}
		\addConstraint{x_{i+1}}{= F(x_i,u_i),}{\quad i=0,\ldots,N-1}
		\addConstraint{x_{i}}{\in \mathcal{X},u_{i}\in \mathcal{U},}{\quad i=0,\ldots,N-1},
	\end{mini}
	and takes the optimal state trajectory of the solution~$(X^*,U^*)$ as output~${X_\mathrm{ref}:=X^*}$ of the {\mpc} algorithm.
	\section{General Method}
\begin{figure}
		\begin{center}
			\vspace{0.15cm}
			\def\svgwidth{.48\textwidth}
			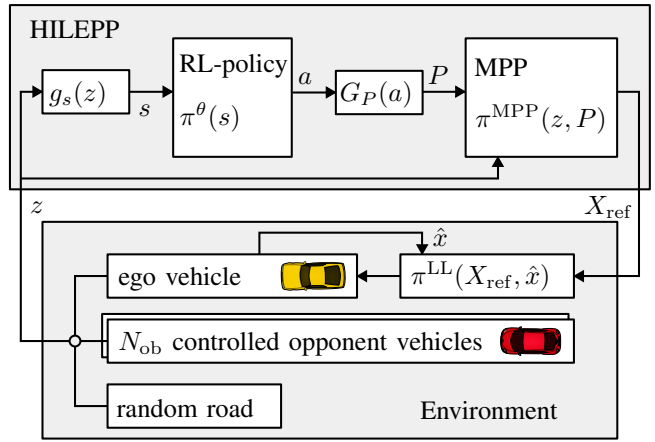
			\caption{Proposed control structure. The multi-vehicle environment constitutes a trajectory tracking ego agent (lowest-level controller~$\pi^\mathrm{LL}(\cdot))$. A state~$z$ concatenates all~$N_\mathrm{ob}+1$ vehicles states and road curvature information. A function~$g_s(z)$ projects the state to a lower dimensional state-space. A {\rl}~$\pi^\mathrm{\theta}(s)$  and an expanding function~$G_P(a)$ modify the cost function of {\mpcfull} ({\mpc}) $\pi^\mathrm{MPC}(z,P)$ by action~$a$. The {\mpc} outputs a feasible and safe trajectory~$X_\mathrm{ref}$ to the ego lowest-level controller.}
			\label{fig:block_diagram}
		\end{center}
	\end{figure}

	We aim at applying our algorithm to a multi-agent vehicle competition on a race-track.
	Our goal is to obtain a planner that performs time-optimal trajectory planning, avoids interactive obstacles and learns strategic behavior, such as blocking other vehicles in simulation.
	We assume fixed policies of the opponents and therefore, do not consider the interaction as a game-theoretical problem \cite{Zhang2021}.
	We use an obstacle avoidance rule, according to the autonomous racing competitions \emph{Roborace}\cite{Roborace} and \emph{F1TENTH} \cite{F1tenth}, where in a dueling situation the following vehicle (FV) is mainly responsible avoiding a crash, however the leading vehicle (LV) must not provoke a crash.
	Unfortunately, to the best of the authors knowledge, there is no rigorous rule for determining the allowed actions of dueling vehicles.
	However, we formalize the competition rules of \emph{F1TENTH} similar to \cite{Li2021}, where the LV only avoids an inevitable crash, which we state detailed in Sec.~\ref{sec:obstacle_constraints}.
	A block diagram of our proposed algorithm is shown in Fig.~\ref{fig:block_diagram}, where we assume a multi-agent environment with a measured state~$z\in \R^{n_z}$, which concatenates the ego agent states~$x$, the obstacle/opponent vehicle states~$x^\mathrm{ob}$ and the road curvature $\kappa(\lon_i)$ on evaluation points $\lon_i$.
	We include prior domain knowledge to get the {\rl} state~$s\in \R^{n_s}$ with the pre-processing function~$s=g_s(z)$.
	For instance, we use relative distances of the opponents to the ego vehicle, instead of absolute values.
	An expansion function~$P=G_P(a)$, with the {\rl}~$a=\pi^\theta(s)$, is used to increase the dimension of the NN output in order to obtain a parametric cost function.
	The expansion function is used to include prior knowledge and to obtain an optimization-friendly cost function in the {\mpc}.
	\section{\mpcfull}
	\label{sec:specific_method}
	Our core component {\mpc} constitutes an MPC formulation that accounts for safety and strong initial racing performance. It comprises a vehicle model, safety constraints and a parameterized cost function, which we will explain in the following section.
	\subsection{Vehicle Model}
	\label{sec:model}
		We use rear-wheel centered kinematic single-track vehicle models in the Frenet coordinate frame, as motivated in previous work \cite{Reiter2021}.
		The models are governed by the longitudinal force~$F_\mathrm{d}$ that accounts for accelerating and braking, and the steering rate~$r$ which is the first derivative of the steering angle~$\delta$.
		The most prominent resistance forces~$F_\mathrm{res}(v)=c_\mathrm{air}v^2+c_\mathrm{roll}\mathrm{sign}(v)$ are included. 
		The air drag depends on the vehicle speed~$v$ with constant~$c_\mathrm{air}$.
		The rolling resistance is proportional to~$\mathrm{sign}(v)$ by the constant~$c_\mathrm{roll}$.
		We drop the sign function, since we only consider positive speed.
		As shown in previous work \cite{Reiter2021, Reiter2021a}, the Frenet transformation~$\mathcal{F}(\cdot)$ relates Cartesian states~$x^\mathrm{C}=[x_\mathrm{e}\quad y_\mathrm{e}\quad \varphi]^\top$, where~$x_\mathrm{e}$ and $y_\mathrm{e}$ are Cartesian positions and $\varphi$ is the heading angle, to the curvilinear states 
		\begin{equation}
		\label{eq:frenet_transformation}
			x^\mathrm{F}=\mathcal{F}(x^\mathrm{C})=[\lon\quad n\quad \alpha]^\top.
		\end{equation}
		The Frenet states are related to the center lane ~${\gamma(\lon)=[\gamma_x(\lon)\quad \gamma_y(\lon)]}$, with signed curvature~$\kappa(\lon)$ and tangent angle~$\varphi^\gamma(\lon)$, where $\lon$ is the 1d-position of the closest point of the center lane, $n$ is the lateral normal distance and~$\alpha$ is the difference of the vehicle heading angle to the tangent of the reference curve.
		Under mild assumptions \cite{Reiter2021}, the Frenet transformation and its inverse 
		\begin{align}
		x^\mathrm{C}=\mathcal{F}^{-1}(x^\mathrm{F})=\begin{bmatrix}
		\gamma_x(\lon)-n\sin(\varphi^\gamma(\lon)) \\
		\gamma_y(\lon)+n\cos(\varphi^\gamma(\lon)) \\
		\varphi^\gamma(\lon)-\alpha 
		\end{bmatrix}
		\end{align}
		are well defined. 
		We use indices "ego" (or no index) and "ob-$n_\mathrm{o}$" to indicate the ego or the $n_\mathrm{o}$-th opponent vehicle and summarize the states~$x = \begin{bmatrix}\lon & n & \alpha & v & \delta \end{bmatrix}^{\top}$ and controls~$u = \begin{bmatrix}F_\mathrm{d} & r\end{bmatrix}^{\top}$. 
		We formulate the Frenet frame vehicle model with parameters for mass~$m$ and length~$l$ as		
		\begin{equation}
		\label{eq:vehicle_model_frenet}
			\dot{x}=
			f(x, u)=
			\begin{bmatrix}
			\frac{v \cos(\alpha)}{1-n \kappa(\lon)}\\
			v \sin(\alpha)\\
			\frac{v}{l}\tan(\delta) - \frac{\kappa(\lon) v \cos(\alpha)}{1-n \kappa(\lon)}\\
			\frac{1}{m}(F_\mathrm{d}-F_\mathrm{res}(v))\\
			r
			\end{bmatrix}.
		\end{equation}
		The discrete states~$x_k$ at sampling time~$k\Delta t$ are obtained by an RK4 integration function $x_{k+1}=F(x_k, u_k, \Delta t)$.
	\subsection{Safety Constraints}
	\label{sec:constraints}
	As stated in Sec.~\ref{sec:general method}, the {\mpc} formulation should restrict trajectories~$X_\mathrm{ref}$ to be within model constraints.
	Since we assume known vehicle parameters and no measurement noise, this can be guaranteed for most limitations in a straight-forward way.
	Nevertheless, the interactive behavior of the opponent vehicles poses a serious challenge to the formulation.
	On one extreme, we could model the other vehicles robustly, which means we account for all possible maneuvers which yields quite conservative constraints.
	On the other extreme, with known parameters of all vehicles, one could model the opponent safety by "leaving space" for at least one possible motion of the opponent without a crash, thus not forcing a crash.
	The later leads to an hard bi-level optimization problem, since the feasibility problem, which is an optimization problem itself, is needed as constraint of the {\mpc}.
	In this work, we aim at a heuristic explained in  Sec.~\ref{sec:obstacle_constraints}.
		\subsubsection{Vehicle Limitations}
		Slack variables~$\sigma = [\sigma_v,\sigma_\alpha,\sigma_n,\sigma_\delta , \sigma_a , \sigma_\mathrm{o}]^{\top}\in \R^6$, with slacks for states $\sigma_v,\sigma_\alpha,\sigma_n,\sigma_\delta$, acceleration $\sigma_a$ and obstacles $\sigma_\mathrm{o}$ are used to achieve numerically robust behavior. 
		We use box constraints for states 
		\begin{subequations}
			\label{eq:static_constraints_x}
			\begin{align} 	
				B_x(\sigma) \coloneqq \big\{ x  \biggm \vert 
				-\sigma_n+\lb{n}\leq &n\leq \ub{n} + \sigma_n,\\
				 -\sigma_\alpha+\lb{\alpha}\leq &\alpha \leq \ub{\alpha}+\sigma_\alpha,\\
				0 \leq & v \leq \ub{v} + \sigma_v,\\
				-\sigma_\delta+\lb{\delta}\leq& \delta \leq \ub{\delta} + \sigma_\delta
				\big\}, 
			\end{align}
		\end{subequations}
		and controls
		\begin{align} 
			\label{eq:static_constraints_u}	
			B_u \coloneqq \big\{ u \bigm\vert 
			\lb{F}_\mathrm{d}\leq F_\mathrm{d} \leq \ub{F}_\mathrm{d},\quad
			\lb{r}\leq r \leq \ub{r}
			\big\}.
		\end{align}
		Further, we use a lateral acceleration constraints set
		\begin{align} 
		\label{eq:static_constraints_acc}
		B_\mathrm{lat}(\sigma) \coloneqq \Bigg\{ x \Bigg|\bigg|\frac{v^2  \tan(\delta)}{l}\bigg| \leq \ub{a}_\mathrm{lat}+\sigma_a \Bigg\},
		\end{align}
		to account for friction limits.
		\subsubsection{Obstacle Constraints}
		\label{sec:obstacle_constraints}
		We approximate the rectangular shape of obstacles in the Cartesian coordinate frame by an ellipse and the ego vehicle by a circle \cite{Nair2022}.
		We assume a predictor of an opponent vehicle~$i$ that outputs the expected Cartesian positions of the vehicle center~${p^{\mathrm{ob}i}_k=[x^{\mathrm{ob}i}_{\mathrm{e},k} \quad y^{\mathrm{ob}i}_{\mathrm{e},k}]^\top}\in \R^2$ with a constraint ellipse shape matrix~$\hat{\Sigma}^{\mathrm{ob}i}_k(x) \in \R^{2\times 2}$ at time step~$k$ that depends on the (Frenet) vehicle state in $x$.
		The ellipse area is increased by $\Sigma^{\mathrm{ob}i}(x) =\hat{\Sigma}^{\mathrm{ob}i}(x) +\Id (r_\mathrm{ego}+\Delta r)^2$ with radii of the ego covering circle~$r_\mathrm{ego}$ and a safety distance $\Delta r$.
		Since the ego vehicle position~${p^\top=[x_\mathrm{e}\quad y_\mathrm{e}]}$ is measured at the rear axis and in order to have a centered covering circle, we project the rear position to the ego vehicle center~$p_\mathrm{mid}$ by
		\begin{equation}
			p_\mathrm{mid}=
			\begin{bmatrix}
			x_\mathrm{e,mid}\\
			y_\mathrm{e,mid}\\
			\end{bmatrix}=P(x^\mathrm{C})=
			\begin{bmatrix}
			x_\mathrm{e}+\frac{l}{2}\cos{\varphi}\\
			y_\mathrm{e}+\frac{l}{2}\sin{\varphi}\\
			\end{bmatrix}
		\end{equation}
		For obstacle avoidance with respect to the ellipse matrix, we use the constraint set in compact notation
		\begin{align}
		\label{eq:obstacle_constraint}
		\begin{split}
			B_{\mathrm{O}}(&x^{\mathrm{ob}},\Sigma^{\mathrm{ob}},\sigma) =\\ 
			&\Bigl\{x\in \R^2\Big|
			\norm{P(\mathcal{F}^{-1}(x))-p^{\mathrm{ob}}}^2_{(\Sigma^{\mathrm{ob}}(x) )^{-1}}
			\geq 1-\sigma_{\mathrm{o}}\Bigr\} .
		\end{split}
		\end{align}
		\subsubsection{Obstacle Prediction}
		\label{sec:obstacle_prediction}
		The opponent prediction uses a simplified model with states $x^\mathrm{ob}=[\lon^\mathrm{ob},n^\mathrm{ob},v^\mathrm{ob}]^\top$ and assumes curvilinear motion depending on the initial estimated state $\hat{x}^\mathrm{ob}$.
		With the constant acceleration force $F^\mathrm{ob}_\mathrm{d}$, the ODE of the opponent estimator can be written as 
		\begin{subequations}
			\label{eq:prediction}
			\begin{align}
				\dot{\lon}^\mathrm{ob}			&= \frac{v^\mathrm{ob}(t) \cos(\hat{\alpha}^\mathrm{ob})}{1-n^\mathrm{ob} \kappa(\lon^\mathrm{ob})}\\
				\dot{n}^\mathrm{ob}			&= v^\mathrm{ob}(t) \sin(\hat{\alpha}^\mathrm{ob})\\
				\dot{v}^\mathrm{ob}			&= 	\frac{1}{m^\mathrm{ob}}F^\mathrm{ob}_\mathrm{d}.
			\end{align}
		\end{subequations}
		Since the FV is responsible for a crash, it \emph{generously} predicts the LV by assuming constant velocity motion, where $F^\mathrm{ob}_\mathrm{d}$ is set to $0$.
		The LV predicts the FV most \emph{insignificantly}, which we realize by assuming a FV full stop with its maximum braking force~$F^\mathrm{ob}_\mathrm{d}=\lb{F}^\mathrm{ob}_\mathrm{d}$.
		In any situation, this allows the FV to plan for at least one safe trajectory (i.e. a full stop), thus the LV does not "provoke" a crash, as required in racing competition rules \cite{Roborace, F1tenth}.
		Besides this minimum safety restrictions, interaction should be learned by the {\rl}.
		We simulate the system forward with a function $\Phi()$, using steps of the RK4 integration function to obtain the predicted states ${[x^\mathrm{ob}_{0},\ldots,x^\mathrm{ob}_{N}]=\Phi(\hat{x}^\mathrm{ob},\hat{\alpha}^\mathrm{ob},F^\mathrm{ob}_\mathrm{d})}$.
		\subsubsection{Recursive feasibility}
		\label{sec:recursive}
		In order to guarantee safety for a finite horizon and constraints \eqref{eq:static_constraints_x}, \eqref{eq:static_constraints_u} and \eqref{eq:static_constraints_acc}, we refer to the concept of recursive feasibility and control invariant sets (CIS) \cite{Rawlings2017}.
		A straight-forward CIS is the trivial set of zero velocity~${\{x \mid  v=0\}}$. 
		An approximation to the CIS, which is theoretically not a CIS but which has shown good performance in practice, is the limited-velocity terminal set~${\mathcal{S}^\mathrm{t} \coloneqq \{x \mid \alpha = 0,  v\leq \ub{v}_\mathrm{max} \}}$.
		For long horizons, the influence of the terminal set vanishes.
	\subsection{Objective}
	\label{sec:objective}
		For the parameterized cost function~$L(X,U,a)$, we propose a formulation with the following  properties:
		\begin{itemize}
			\item Simple structure for reliable and fast NLP iterations
			\item Expressive behavior of the vehicle related to the strategic driving application
			\item Low dimensional action vector~$a$
			\item Good initial performance with constant actions
		\end{itemize}
		The first property is achieved by restricting the cost function to a quadratic form.
		The second property is achieved by formulating the state reference in the Frenet coordinate frame.
		The final properties of a low dimensional action space and a good initial performance are achieved by interpreting the actions as reference lateral position~$n_\mathrm{ref}$ and reference speed~$v_\mathrm{ref}$.
		By setting the reference speed, also the corresponding longitudinal state~$\lon_{\mathrm{ref},k}$ of a curvilinear trajectory is defined by $\lon_{\mathrm{ref},k}=\hat{\lon}+k\Delta t v_\mathrm{ref}$.
		The reference heading angle miss-match~$\alpha_\mathrm{ref}$ as well as the steering angle~$\delta_\mathrm{ref}$ are set to zero, with fixed weights~$w_\alpha$ and~$w_\delta$, since these weights are tuned for a smooth driving behavior.
		Setting the reference speed~$v_{\mathrm{ref}}$ above maximum speed corresponds to time-optimal driving \cite{Kloeser2020}.
		We compare the influence using references~$v_\mathrm{ref}, n_\mathrm{ref}$ with their associated weights~$w_v, w_n$ ({\mpcrl-II}) in the action space or with fixed weights without using them in the action space ({\mpcrl-I}).
		We use the action-dependent stage cost matrix~$Q_{\mathrm{w}}(a)$ with ${Q_{\mathrm{w}}:\R^{n_a}\rightarrow \R^{n_x \times nx}}$ and a cost independent terminal cost~$Q^t\in \R^{n_x \times nx}$.
		We set the values of $R,$ $Q_0$ and $Q^\mathrm{t}$ to values that correspond to driving smoothly and time-optimal.
		With constant action inputs~$a=\bar{a}$, this leads to a strong initial performance in the beginning of training the {\rl}.
		With the action-dependent reference values~$\xi_{\mathrm{ref},k}(a)\in\R^{n_x}$, we can write the expanding function as
		\begin{align}
				G_P(a):a \rightarrow \Big(\xi_{\mathrm{ref},0}(a),\ldots,\xi_{\mathrm{ref},N}(a),Q_{\mathrm{w}}(a)\Big)
		\end{align}
		which maps $n_a$ to $n_x^2(N+1)+n_x(N+1)$ dimensions for cost matrices and reference values.
	\subsection{NLP formulation}
	\label{sec:nlp}
		We state the final NLP, using the vehicle model \eqref{eq:vehicle_model_frenet}, the MPC path constraints for obstacle avoidance \eqref{eq:obstacle_constraint}, vehicle limitations \eqref{eq:static_constraints_x},  \eqref{eq:static_constraints_u} and \eqref{eq:static_constraints_acc} and the parametric cost functions of \eqref{eq:leremo_filter}.
		The full objective, including slack variables~${\Xi=[\sigma_0,\ldots,\sigma_N]\in\R^{6\times N}}$ for each stage, associated L2 weights $Q_{\sigma,2}=\textrm{diag}(q_{\sigma,2})\in \R^{6 \times 6}$ and L1 weights $q_{\sigma,1} \in \R^6$, reads as
		\begin{align}
		\begin{split}
			L(&X,U,a,\Xi) = \sum_{k=0}^{N-1} \norm{x_k-\xi_{\mathrm{ref},k}(a)}_{Q_{\mathrm{w}}(a)}^2 + \norm{u_k}_{R}^2 \\&  +
			\norm{x_N-\xi_{\mathrm{ref},N}(a)}_{Q^t}^2+ \sum_{k=0}^{N}\norm{\sigma_k}_{Q_{\sigma,2}}^2 + |q_{\sigma,1}^\top \sigma_k|.
		\end{split}
		\end{align}
		Together with the predictor for time step~$k$ of the $j$-th future opponent vehicle states, represented as bounding ellipses with the parameters~$p^{\mathrm{ob},j}_i,\Sigma^{\mathrm{ob},j}_i$, the parametric NLP can be written as
		\begin{mini}
			{X,U,\Xi}
			{L(X,U,a,\Xi)}			
			{\label{eq:final_mpc_policy}} 
			{} 
			\addConstraint{x_0}{= \hat{x}, \quad \Xi \geq 0, \quad x_N \in \mathcal{S}^\mathrm{t}}{}
			\addConstraint{x_{i+1}}{= F(x_i,u_i),}{\quad i=0,\ldots,N-1}
			\addConstraint{\state_{i}}{\in B_x(\sigma_k)\cap B_\mathrm{lat}(\sigma_k)}{\quad i=0,\ldots,N}
			\addConstraint{\state_{i}}{\in B_{\mathrm{ob}}(p^{\mathrm{ob},j}_i,\Sigma^{\mathrm{ob},j}_i,\sigma_k)}{\quad i=0,\ldots,N}
			\addConstraint{}{}{\quad j=0,\ldots,N_\mathrm{ob}}
			\addConstraint{\controls_i}{\in B_u,}{\quad i=0,\ldots,N-1}.
		\end{mini}
		The final {\mpc} algorithm is stated in Alg.~\eqref{alg:ppp}.
		\begin{algorithm}
			\label{alg:ppp}
			\SetKwData{Left}{left}\SetKwData{This}{this}\SetKwData{Up}{up}
			\SetKwFunction{Union}{Union}\SetKwFunction{FindCompress}{FindCompress}
			\SetKwInOut{Input}{input}\SetKwInOut{Output}{output}
			\Input{action~$a$, ego states~$\hat{x}$, $N_\mathrm{ob}$ obstacle states~$\hat{x}^{\mathrm{ob}}$}
			\Output{planned trajectory $X_\mathrm{ref}$}
			\For{j in range($N_\mathrm{ob}$)}{
				\If{$\hat{\lon}^\mathrm{ob}\leq\hat{\lon}$}
				{Consider opp. as FV: $F^\mathrm{ob}_\mathrm{d}\gets \lb{F}^\mathrm{ob}_\mathrm{d}$} 
				\Else
				{Consider opp. as LV $F^\mathrm{ob}_\mathrm{d}\gets0$}
				{Predict $ [x^\mathrm{ob}_{0},\ldots,x^\mathrm{ob}_{N}]=\Phi(\hat{x}^\mathrm{ob},\hat{\alpha}^\mathrm{ob},F^\mathrm{ob}_\mathrm{d})$}\;
				{Compute constraint ellipses $\Sigma^{\mathrm{ob},j}_k=\Sigma_0(\varphi^{\mathrm{ob},j})$}\;
			}
			Compute weights $\big(\zeta_{\mathrm{ref},k},Q_{\mathrm{w}}\big)\gets G_P(a)$\;
			{$X_\mathrm{ref}\gets$Solve NLP \eqref{eq:leremo_filter} with $\big(\zeta_{\mathrm{ref},k},Q_{\mathrm{w}}\big)$}\;
			\caption{{\mpc}}
		\end{algorithm}
\section{{\mpcrldef}}
\label{sec:rl_policy}
The {\mpc} of Sec.~\ref{sec:specific_method} plans safe and time-optimal, but not strategically.
Therefore, we learn a policy~$\pi^\theta$ with RL that decides at each time step how to parameterize the {\mpc} to achieve a strategic goal.
Since we assume stationary opponent policies, we can apply standard, i.e. single-agent RL algorithms \cite{Zhang2021} and solve for a best response. 
In the following, we give a short theoretical background to policy gradient methods and then describe the training procedure in detail.
\subsection{Policy gradient}
RL requires a Markov Decision Process (MDP) framework.
A MDP consists of a state space~$\mathcal{S}$, an action space~$\mathcal{A}$, a transition kernel~$P(s'  \mid s, a)$, a reward function~$R: \mathcal{S} \times \mathcal{A} \mapsto \R$ that describes how desirable a state is (equal to the negative cost) and a discount factor~$\gamma \in [0, 1)$.
The goal for a given MDP is finding a policy~$\pi^\theta: \mathcal{S} \mapsto \mathcal{A}$ that maximizes the expected discounted return
\begin{equation}
\label{eq:mdp}
\begin{split}
J(\pi^\theta) = \mathbb{E}\left[\sum_{t=0}^\infty \gamma^t R(s_t, a_t)\mid s_0 = s \right],\\
s_t \sim P(s_{t+1} \mid s_t, a_t), a_t \sim \pi^\theta(s_t).
\end{split}
\end{equation}
where $s_t$ is the state and $a_t$ the action taken by the policy~$\pi^\theta$ at time step~$t$.
An important additional concept is the state-action value function
\begin{equation}
\label{eq:state-action}
Q^{\pi^\theta}(s, a) = \mathbb{E}\left[\sum_{t=0}^\infty \gamma^t R(s_t, a_t)\mid s_0 = s, a_0=a \right]
\end{equation}
that is the expected value of the policy~$\pi^\theta$ starting in state~$s$ and taking an action~$a$.
In general, it is not possible to find an optimal policy~$\pi^\theta$ by directly optimizing $\theta$ in \eqref{eq:mdp}.
The expectation in \eqref{eq:mdp} might be computationally intractable or the exact transition probabilities~$P$ may be unknown.
Thus, the policy gradient~$\nabla J(\pi^\theta)$ is approximated by only using transition samples from the environment.
We sample these transitions from a simulator, however, they could also come from real-world experiments.
A particularly successful branch of policy gradient methods are \emph{actor-critic} methods \cite{sutton2018reinforcement}, where we train two neural networks, an actor~$\pi^\theta$ and a critic~$Q^\phi $.
The critic estimates the value of a chosen action and is trained by minimizing the temporal difference loss
\begin{equation}
\label{eq:critic}
\begin{split}
J&_Q(\phi)  =\\ &\mathbb{E}_{s, a, r, s' \sim \mathcal{D}}\left[ \left( r_t + \gamma Q_{\phi'} \left(s_{t+1}, \pi^\theta(s_{t+1})\right) - Q_{\phi} (s_{t}) \right)^2 \right].
\end{split}
\end{equation}
The trained critic is used to train the actor with the objective
\begin{equation}
\label{eq:actor}
\begin{split}
J_\pi(\theta) = \mathbb{E}_{s, a, r, s' \sim \mathcal{D}}\left[ Q^\phi (s_t, \pi^\theta (s_t)) \right].
\end{split}
\end{equation}
To derive the gradient for the policy~$\pi^\theta$ from \eqref{eq:actor} we can use the chain rule \cite{silver2014deterministic} 
\begin{equation}
\label{eq:dpgrad}
\nabla_\theta J_\pi(\theta) = \\ \mathbb{E}_{s, a, r, s' \sim \mathcal{D}}\left[ \nabla_\theta \pi^\theta (s) \nabla_a Q^\phi (s, a) \right].
\end{equation}
The soft-actor critic method, introduced by \cite{haarnoja2018soft} enhances the actor-critic method by adding an additional entropy term in to \eqref{eq:actor} and using the \emph{reparameterization trick} to calculate the gradient. 
For a complete description we refer to \cite{haarnoja2018soft}.
\subsection{Training Environment}
We reduce the RL state-space, based on domain knowledge which we put into the function $g_s(z_k)$. 
The race track layout is approximated by finite curvature evaluations~${\kappa(\lon+d_i)}$ at different longitudinal distances~$d_i$ relative to the ego vehicle position~$\lon$, for $i=1,\ldots,N_\kappa$.
For the RL ego state~$s_\mathrm{ego}(z_k)=[n,v,\alpha]^\top$, we include the lateral position $n$, the velocity $v$ and the heading angle miss-match $\alpha$.
For opponent~$i$, we additionally add the opponent longitudinal distance ${\lon_{\textrm{ob}}-\lon}$ to the ego vehicle to state~$s_{\textrm{ob}_i} =[\lon_{\textrm{ob}_i}-\lon,n_{\textrm{ob}_i},v_{\textrm{ob}_i}, \alpha_{\textrm{ob}_i}]^\top$.
Combined, we get the following state definition for the RL agent
\begin{align}
	\begin{split}
	s_{k} &= g_s(z_k) =	\\
	&[
	\kappa(\lon+d_i),
	\ldots ,
	\kappa(\lon+d_N),
	s_\mathrm{ego}^\top,
	s_{\textrm{ob}_1}^\top ,
	\ldots ,
	s^\top_{\textrm{ob}_{N_\textrm{ob}}}]^\top.
	\end{split}
\end{align}
We propose a simple reward that encourages time-optimal and strategic driving:
For driving time-optimal, we reward the progress on the race track by measuring the velocity of the ego vehicle projected point on the center line~$\dot{s}_k$.
For driving strategically, we reward ego vehicle overall rank.
Combined, we get the reward function
\begin{align}
	\label{eq:reward}
    R(s, a) = \frac{\dot{s}}{200} + \sum_{i=1}^{N_\mathrm{ob}} 1_{\lon_k > \lon_k^{\mathrm{ob}_i}}.
\end{align}
At each time step the high-level RL policy chooses a parameter for the {\mpc}, thus the action space is the parameter space of the {\mpc}.
	For training the {\rl}, an essential part is the simulation function of the environment~$z_\mathrm{next}=\mathrm{sim}(z,\kappa(\cdot))$, which we simulate for $n_\mathrm{epi}$ episodes and a maximum of $n_\mathrm{scene}$ steps. 
	The road layout defined by $\kappa(\lon)$ is randomized within an interval~$[\shortminus0.04,0.04]\mathrm{m}^{-1}$ before each training episode. 
	The curvature is set together with initial random vehicle states~$z$ by a reset function~$(z,\kappa(\lon))=Z()$.
	We use Alg.~\ref{alg:alg_learning} for training and Alg.~\ref{alg:alg_deployment} for the final deployment of {\mpcrl}.
	\begin{algorithm}
	\label{alg:alg_learning}
		\SetKwData{Left}{left}\SetKwData{This}{this}\SetKwData{Up}{up}
		\SetKwFunction{Union}{Union}\SetKwFunction{FindCompress}{FindCompress}
		\SetKwInOut{Input}{input}\SetKwInOut{Output}{output}
		\Input{number of episodes $n_\mathrm{epi}$, maximum scenario steps $n_\mathrm{scene}$, reset function $(z,\kappa(\lon))=Z()$, reward function $r(z)$}
		\Output{learned policy $\pi^\theta(\lon)$}
		\For{j in range($n_\mathrm{epi}$)}{
			{reset+randomize environment $(z,\kappa(\lon)) \gets Z()$}\;
			\For{i in range($n_\mathrm{scene}$)}{
				get NN input state $s\gets g_s(z)$\;
				{get high-level action $a\gets\pi^\theta(s)$}\;
				{evaluate planner $X_\mathrm{ref}\gets$\mpc$(a,z)$}\;
				simulate environment $z_\mathrm{next}=\mathrm{sim}(X_\mathrm{ref})$\;
				get reward $(r,\mathrm{done})\gets R(z_\mathrm{next},a)$\;
				RL update $\theta\gets$train$(z,z_\mathrm{next},r,a)$\;
				\If{done}{exit loop}
				z$\gets z_\mathrm{next}$
			}
		}
		return $\pi^\theta(s)$\;
		\caption{{\mpcrl} training}
	\end{algorithm}
	\begin{algorithm}
	\label{alg:alg_deployment}
		\SetKwData{Left}{left}\SetKwData{This}{this}\SetKwData{Up}{up}
		\SetKwFunction{Union}{Union}\SetKwFunction{FindCompress}{FindCompress}
		\SetKwInOut{Input}{input}\SetKwInOut{Output}{output}
		\Input{environment state $z$, trained policy $\pi^\theta(s)$}
		\Output{reference trajectory $X_\mathrm{ref}$}
		compute NN input state $s\gets g_s(z)$\;
		compute {\rl} output $a\gets \pi^\theta(s)$\;
		return {\mpc} output $X_\mathrm{ref}\gets$\mpc$(z,a)$\;
		\caption{{\mpcrl} deployment}
	\end{algorithm}
\section{Simulated Experiments}
\label{sec:results}	
\begin{figure}
		\begin{center}
			\def\svgwidth{.48\textwidth}
			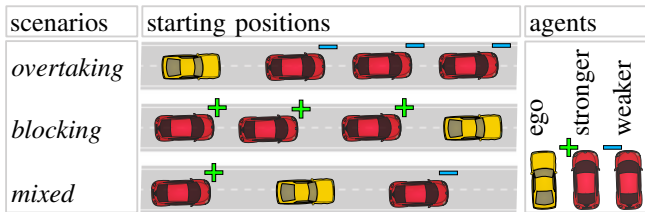
			\caption{Scenarios differ in the initial rank and performance of vehicles.}
			\label{fig:scenarios}
		\end{center}
	\end{figure}

	We evaluate (Alg.~\ref{alg:alg_deployment}) and train (Alg.~\ref{alg:alg_learning}) {\mpcrl} on three different scenarios that resemble racing situations (cf. Fig.~\ref{fig:scenarios}).
	The first scenario \emph{overtaking} constitutes three "weaker", initially leading opponent agents, where "weaker" relates to the parameters of maximum accelerations, maximum torques and vehicle mass (cf. Tab.~\ref{table:vehicle_scenario_comparison}).
	The second scenario \emph{blocking} constitutes three "stronger", initially subsequent opponents.
	The ego agent starts in between a stronger and a weaker opponent in a third \emph{mixed} scenario.
	Each scenario is simulated for one minute, where the ego agent has to perform best related to the reward \eqref{eq:reward}.
	We train the {\mpcrl} agent with the different proposed action interfaces (I: $\mathcal{A}=\{n_\mathrm{ref}, v_\mathrm{ref}\}$, II: $\mathcal{A}=\{n_\mathrm{ref}, v_\mathrm{ref},w_n, w_v\}$).
	Opponent agents, as well as the ego agent base-line, are simulated with the state-of-the-art {\mpc} (Alg.~\ref{alg:ppp}) with a fixed action~$a$ that accounts for non-strategic time-optimal driving with obstacle avoidance.
	\begin{table}
	\centering
	\ra{1.1}
	\begin{tabular}{@{}lllll@{}}
		\addlinespace
		\toprule
		Name & Variable & Ego & "Weak" & "Strong"  \\
		 &  & Agent & Agent & Agent  \\
		\midrule
		wheelbase & $l_\mathrm{r}, l_\mathrm{f}$ & $1.7$   & $1.7$&$1.7$\\
		chassis lengths& $l_\mathrm{r,ch}, l_\mathrm{f,ch}$& $2$ &$2$  &$2$ \\
		chassis width & $w_\mathrm{ch}$& $1.9$& $1.9$&$1.9$\\
		mass & m & 1160& 2000&600\\
		max. lateral acc. & $\lb{a}_\mathrm{lat},\ub{a}_\mathrm{lat}$ & $\pm 8$& $\pm 5$&$\pm 13$\\
		max. acc. force& $ \ub{F}_\mathrm{d}$ & $10\mathrm{kN}$ &$8\mathrm{kN}$ &$12\mathrm{kN}$\\
		max. brake force& $\lb{F}_\mathrm{d}$ & $20\mathrm{kN}$ & $20\mathrm{kN}$&$20\mathrm{kN}$\\
		max. steering rate& $\lb{r},\ub{r}$ & $\pm0.39$ &$\pm0.39$ &$\pm0.39$\\
		velocity bound & $\ub{v}$ & $60$& $60$&$60$\\
		steering angle bound & $\lb{\delta},\ub{\delta}$ & $\pm0.3 $ &$\pm0.3 $ &$\pm0.3 $\\
		road bounds & $\lb{n},\ub{n}$ & $\pm7 $ &$\pm7 $ &$\pm7 $\\
		\bottomrule
	\end{tabular}
	\caption{Vehicle model parameters. SI-units, if not stated explicitly.}
	\label{table:vehicle_scenario_comparison}
	\end{table}
	We perform hyper-parameter (HP) search for the RL parameters with Hydra \cite{Yadan2019Hydra} and its \emph{Optuna} HP optimizer extension by \cite{akiba2019optuna}.
	The search space was defined by $[10^{-5}, 10^{-3}]$ for the learning rate, $\tau \in [10^{-5}, 10^{-2}]$ for the polyak averaging of the target networks, $\{64, 128, 256\}$ for the width of the hidden layers, $\{1, 2, 3\}$ for the number of hidden layers and $\{128, 256\}$ for the batch size.
	We used the average return of $30$ evaluation episodes after training for $10^5$ steps as the search metric.
	We trained on randomized scenarios for $5\cdot 10^5$ steps with $10$ different seeds on each scenario.
	For estimating the performance of the final policy, we evaluated the episode return (sum of rewards) on $100$ episodes.
	We further compare our trained {\mpcrl} against a pure RL policy that directly outputs the controls.
	The final experiments where run on a computing cluster where all $30$ runs for one method where run on 8 GeForce RTX 2080 Ti with a AMD EPYC 7502 32-Core Processor with a training time of around 6 hours.
	We use the NLP solver \texttt{acados}~\cite{Verschueren2021} with \texttt{HPIPM}~\cite{Frison2020a}, RTI iterations and a partial condensing horizon of $\frac{N}{2}$.
	\begin{table}
		\vspace{0.15cm}
		\centering
		\ra{1.2}
		\begin{tabular}{@{}lll@{}}
			\toprule
			Name & Variable & Value  \\
			\midrule
			nodes / disc. time & $N$/ $\Delta t$  & $50$/ $0.1$\\
			terminal velocity & $\ub{v}_N$ & $15$ \\
			state weights & $q$ & $[1, 500, 10^3, 10^3, 10^4]\Delta t$\\
			terminal state weights  & $q_N$ & $[10, 90, 100, 10, 10]$\\
			L2 slack weights & $q_{\sigma,2}$ & $[10^2, 10^3, 10^6, 10^3, 10^6, 10^6]$\\
			L1 slack weights  & $q_{\sigma,1}$ & $[0, 0, 10^6, 10^4, 10^7, 10^6]$\\
			control weights  & $R$ & diag($[10^{-3}, 2\cdot 10^6])\Delta t$\\
			\bottomrule
		\end{tabular}
		\caption{Parameters for {\mpc} in SI units}
		\label{table:parameters_mpc}
	\end{table}
	\subsection{Results}
\begin{figure}
		\vspace{0.15cm}
		\begin{center}
			\includegraphics[width=0.45\textwidth,trim={0.2cm 0.2cm 0.2cm 0.2cm},clip]{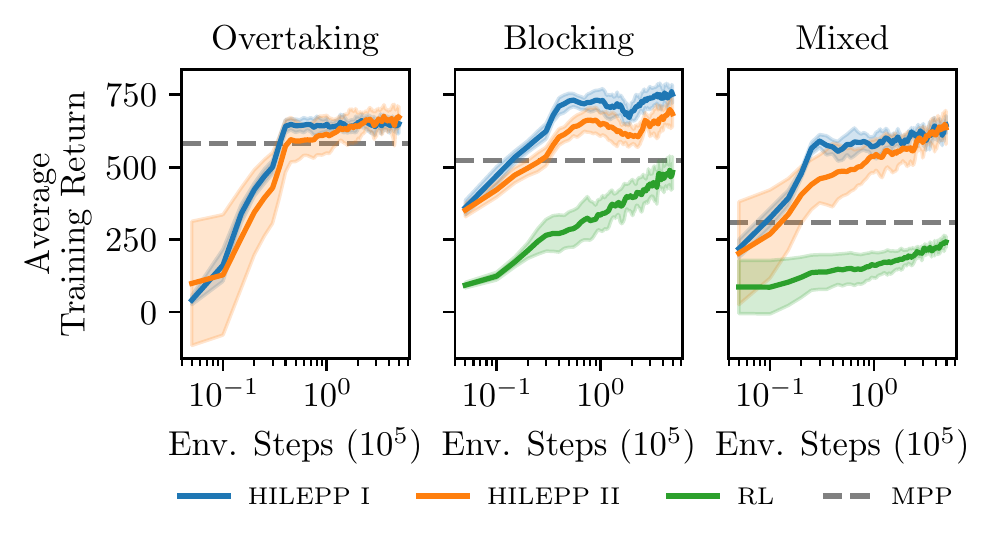}
			\caption{Training performance of average episode return (sum of rewards) of {\mpcrl} with different action interfaces (I: actions $v_\mathrm{ref}, n_\mathrm{ref}$, II: actions $v_\mathrm{ref}, n_\mathrm{ref}, w_v, w_n$), pure RL and the {\mpc} baseline. We used a moving average over $1000$ steps. The confidence intervals are given by the standard deviation.}
			\label{fig:training}
		\end{center}
	\end{figure}

\begin{figure}
		\vspace{0.15cm}
		\begin{center}
			\includegraphics[width=0.4\textwidth,trim={0.2cm 0.2cm 0.2cm 0.2cm},clip]{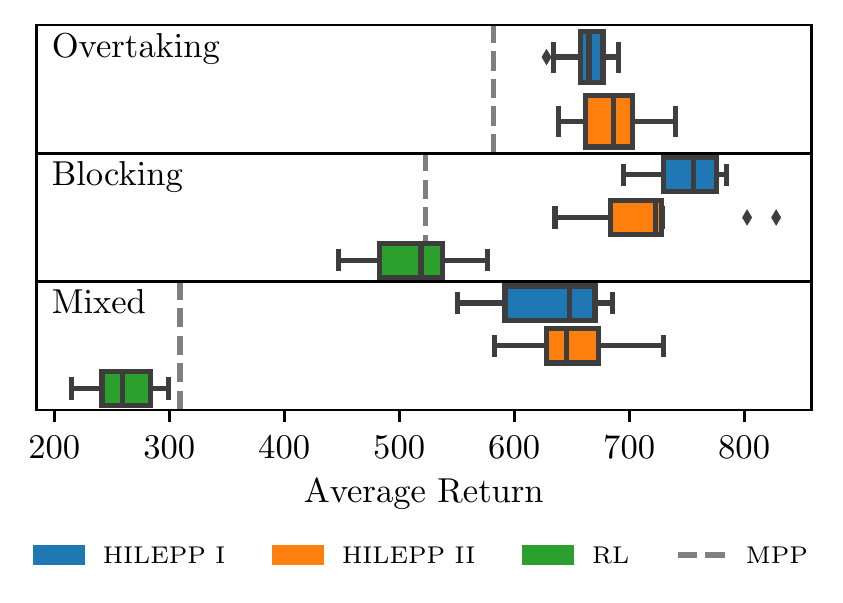}
			\caption{Final episode return of 100 evaluation runs of the proposed interfaces for different scenarios (Fig.~\ref{fig:scenarios}).}
			\label{fig:performance}
		\end{center}
	\end{figure}

\begin{figure}
		\begin{center}
			\includegraphics[width=0.48\textwidth,trim={0.2cm 0.2cm 0.2cm 0.2cm},clip]{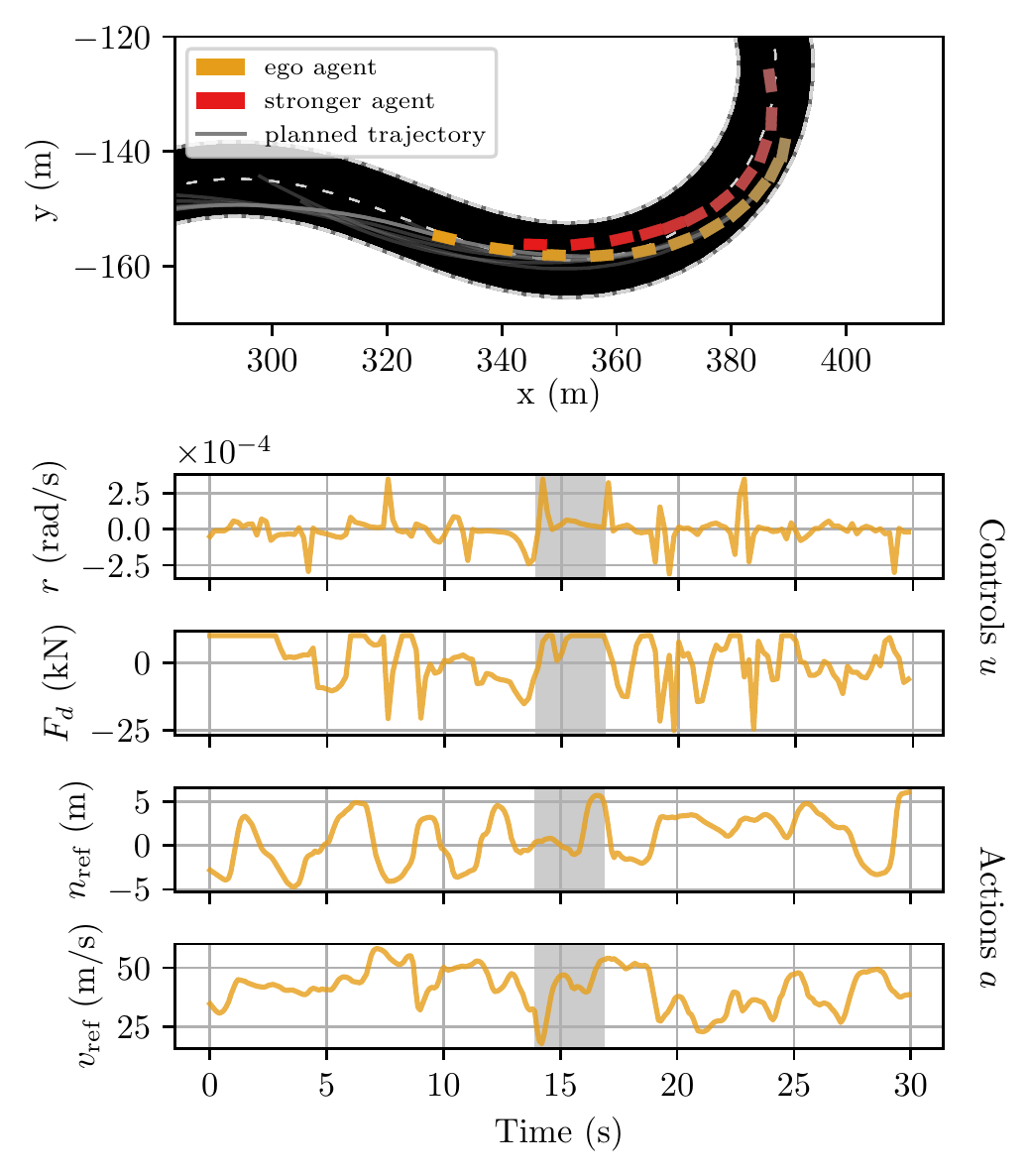}
			\caption{Exemplary evaluation episode of the \mpcrl-I planner in the mixed scenario. On the bottom, the controls of the {\mpc} and the actions of the {\rl} are shown. The grey box shows a time window, where snapshots of a blocking maneuver are shown in the top plot.}
			\label{fig:episode_plot}
		\end{center}
	\end{figure}

	In Fig.~\ref{fig:training}, we compare the training performance related to the reward \eqref{eq:reward} and in Fig.~\ref{fig:performance}, we show the final performance of the two {\mpcrl} formulations.
	{\mpcrl} quickly outperforms the base-line {\mpc} as well as the pure RL formulation.
	With a smaller action space, {\mpcrl-I} learns faster, however, with more samples, {\mpcrl-II} outperforms the smaller action space in two out of three scenarios (\emph{mixed} and \emph{overtaking}).
	Despite using state-of-the-art RL learning algorithms and an extensive HP search on GPU clusters, the pure RL agent can not even outperform the {\mpc} baseline.
    Furthermore, the pure RL policy could not prevent crashes, whereas {\mpc} successfully filters the actions within {\mpcrl} to safe actions that do not cause safety violations.
	Notably, due to the struggle of the pure RL agent with lateral acceleration constraints, it has learned a less efficient strategy to drive slowly and just focus on blocking subsequent opponents in scenarios \emph{blocking} and \emph{mixed}.
	Therefore, pure RL could not perform efficient overtaking maneuvers in the \emph{overtaking} scenario and yields insignificant returns (consequently excluded in Fig.~\ref{fig:performance}). 
	In Tab.~\ref{tab:timings} we show, that {\mpcrl} is capable of planning trajectories with approximately $100$Hz, which is sufficient and competitive for automotive motion planning \cite{betz_2022_1}.
	A rendered plot of learned blocking is shown in Fig.~\ref{fig:episode_plot}, where also the time signals are shown of how the {\rl} sets the references of {\mpcrl-I}.
	A rendered simulation for all three scenarios can be found on the website \url{https://rudolfreiter.github.io/hilepp_vis/}.
	\begin{table}
		\centering
		\ra{1.2}
		\begin{tabular}{@{}lll|lll@{}}
			\toprule
			Module 		& Mean$\pm$ Std. & Max &  Module & Mean$\pm$ Std. & Max  \\
			\midrule
			{\mpc} 		&$ 5.45 \pm 2.73$ &  $8.62$&{\mpcrl}-I	& $6.90\pm 3.17$&  $9.56$\\
			{RL policy} & $0.13 \pm 0.01$&  $0.26$&{\mpcrl}-II	& $7.41\pm 2.28$&  $9.21$\\
			\bottomrule
		\end{tabular}
		\caption{Computation times (ms) of modules.}
		\label{tab:timings}
	\end{table}

\section{Conclusions}
\label{sec:conclusion}
	We have shown a hierarchical planning algorithm for strategic racing.
	We use RL to train for strategies in simulated environments and have shown to outperform a basic time-optimal and obstacle avoiding approach, as well as pure deep-learning based RL in several scenarios.
	The major drawback of our approach is the restrictive prediction and the stationary policy of opponents.
	More involved considerations would need to use multi-agent RL (MARL) based on Markov games, which is a challenging research area with still many open problems \cite{Zhang2021}.
	Both, an interactive prediction and MARL will be considered in future work.
\section*{ACKNOWLEDGMENT}
This research was supported by DFG via Research Unit FOR 2401 and project 424107692 and by the EU via ELO-X 953348.
The authors thank Christian Leininger, the SYSCOP and the Autonomous Racing Graz team for inspiring discussions.
\bibliographystyle{IEEEtran} 
\bibliography{IEEEabrv,syscop,library2}
\end{document}